\documentclass[10pt,twocolumn,letterpaper]{article}

\usepackage{iccv}
\usepackage{times}
\usepackage{epsfig}
\usepackage{graphicx}
\usepackage{amsmath}
\usepackage{amssymb}
\usepackage{soul}
\usepackage{rotating}
\usepackage{commath}
\usepackage{textcomp}
\usepackage[linesnumbered,ruled]{algorithm2e}
\usepackage{multirow}
\usepackage{placeins}
\usepackage{caption}
\usepackage{subcaption}
\DeclareMathOperator*{\argmax}{arg\,max}

\usepackage[breaklinks=true,bookmarks=false]{hyperref}

\iccvfinalcopy 

\def\iccvPaperID{****} 

\begin{document}

\title{RECAL: Reuse of Established \emph{CNN} classifer Apropos
	unsupervised Learning paradigm }

\author{Jayasree Saha\\
{\tt\small jayasree.saha@iitkgp.ac.in}
\and
Jayanta Mukhopadhyay\\
{\tt\small jay@cse.iitkgp.ac.in}
}

\maketitle

\begin{abstract}
 Recently, clustering with deep network framework has attracted attention of several researchers in the computer vision community. Deep framework gains extensive attention due to its efficiency and scalability towards large-scale and high-dimensional data. 
 In this paper, we transform supervised \emph{CNN} classifier architecture into an unsupervised clustering model, called {{\emph RECAL}}, which jointly learns discriminative embedding subspace and  cluster labels. {\emph RECAL} is made up of feature extraction layers which are convolutional, followed by unsupervised classifier layers which is fully connected. A multinomial logistic regression function ({\emph softmax}) stacked
 on top of classifier layers. We train this network using stochastic gradient descent ({\emph SGD}) optimizer. However, the successful implementation of our model is revolved around the design of loss function.  Our loss function uses the heuristics that true partitioning entails lower entropy given that the class distribution is not heavily skewed. This is a trade-off between the situations of ``skewed distribution" and ``low-entropy". To handle this, we have proposed classification entropy and class entropy which are the two components of our loss function. In this approach, size of the mini-batch should be kept high.
 Experimental results indicate the consistent and competitive behavior of our model for clustering well-known digit, multi-viewed object and face datasets. Morever, we use this model to generate unsupervised patch segmentation for multi-spectral \emph{LISS-IV}
 images. We observe that it is able to distinguish built-up area, wet land, vegetation and  water-body from the underlying scene.
 
\end{abstract}

\section{Introduction}
Clustering of large scale data is an active area of research in deep learning and big-data domain. It has gained significant attention as labeling data is costly and anomalous in real-world data. 
For example, getting ground truths for satellite imagery is difficult. A few disaster management applications suffer from that. Nevertheless, clustering algorithms have been studied widely. But, its performance drops drastically when it encounters high dimensional data. Also, their time-complexity increases dramatically. To handle curse of dimensionality, original data space is projected into low dimensional manifold and clustering is applied using them with standard algorithms on that new manifold. However, this approach fails when number of samples reaches to millions or billions. 
To cope up with the situation, several deep architecture based clustering approaches have been proposed in recent years. Most of them are \emph{CNN} or \emph{feed-forward} based auto encoder architecture~\cite{DEC2015,DEPICT,JULE2016,DCAE_clustering}, embedded with simple, \emph{k-means} like, clustering algorithm. These models have the tendency of learning features of the dataset and their corresponding cluster labels jointly. They employ a heuristic for clustering and use it along with the reconstruction loss function for training the model. Therefore, the bottleneck is to get initial clusters to accommodate clustering objective function in the loss function. The intuitive choice for the above problem  is to train well-known auto-encoder model with reconstruction loss to generate new embedding subspace. Then, best known clustering algorithm is employed on that subspace to generate clusters~\cite{DEPICT}. Thereafter, new model is ready to train using two components of the loss function. However, another choice is to employ \emph{k-means} like algorithm to get initial cluster labels.
Other than auto-encoders, there are stacked convolutional layer based deep architecture exist~\cite{CCNN2018}. 
However, they pre-trained their model with large-scale \emph{ImageNet} dataset in order to use clustering objective function.~
It is to be noted that if the model architecture is similar to well-known classifiers architecture like \emph{AlexNet}~\cite{Alexnet}, \emph{VGG}~\cite{VGGnet}, \emph{ResNet}~\cite{resnet}, \emph{DenseNet}~\cite{denseNet} then learned parameters on those layers are eventually transferred. 
However, pre-training is only an approach of initializing the model parameters as this can not fit the data distribution of the dataset under consideration.
Also, it is speeding up the convergence of iterations for real time training.
To put it in a nutshell, the existing unsupervised deep approaches have following pitfalls: 1) training the network from scratch, and 2) obtaining initial cluster labels.
To address the above mentioned pitfalls, we propose a deep clustering algorithm which does not require any initial cluster labels. 
In this paper, we consider the following two issues:
\begin{enumerate}
	\item Can we utilize features generated from well-known pre-trained deep networks to develop clustering algorithm such that computational cost and memory requirement be reduced?
	
	\item Can we perform clustering of these feature vectors under the same deep learning framework in absence of any information on labeling of these data?
\end{enumerate}

In recent years, \emph{AlexNet}~\cite{Alexnet}, \emph{VGG}~\cite{VGGnet}, \emph{ResNet}~\cite{resnet}, \emph{DenseNet}~\cite{denseNet} receive high popularity in \emph{ImageNet} challenge. Thus, it motivates us to choose one of the networks which have the capability of extracting good feature representatives. However, these networks are supervised. Hence, they can not be used in the scenario where no label is present.  In this paper, we propose a loss function which enable the unsupervised labeling of data using these networks. For the purpose of illustration, we have chosen typically \emph{VGG-16} with batch normalization in our experiments. Also, we replace all fully connected and following layers by a single fully connected layer followed by batch normalization and \emph{ReLU} layers. Our loss function is formulated using following entropy based clustering heuristics:
\begin{enumerate}
	\item For a true partitioning, low entropy is desirable. This demands highly skewed probability distribution for a sample to be assigned to a cluster.
	
	\item Any random partitioning than the actual one should lead to higher entropy, given the class distribution is not heavily skewed. For example, when  all  the feature vectors form trivially a single cluster, the distribution becomes highly skewed. In this case, though the entropy is zero, it does not reflect any meaningful information regarding partitioning.
\end{enumerate}
There is a trade-off between these two factors. We propose a loss function for the network which takes into account these two factors for a given number of clusters.
To summarize, the major contributions of our work
are:
\begin{enumerate}
	\item We propose a simple but effective approach to reuse well-known supervised deep architectures in an unsupervised learning paradigm.
	\item We formulate an entropy based loss function to guide clustering and deep representation learning in a single framework that can be
	transferred to other tasks and datasets.
	\item Our experimental results show that the proposed framework is consistent and competitive irrespective of different datasets.
	\item Finally, we use this network in remote sensing domain to provide  segmentation for multi-spectral images.
\end{enumerate}

\section{Approach}
\subsection{Notation}
We represent an image set with $n$ images by $X=\{x_{1}, \cdots, x_{n}\}$ where each sample $x_{i}$ represents an image of resolution $n_{1} \times n_{2}$. The clustering task is to group image set $X$ into $K$ clusters $Y=\{y_{1}, \cdots y_{K}\}$. Using embedding function $f_{\theta}:X \rightarrow Z$, we transform raw image samples $X$ into its embedding subspace $Z=\{z_{1}, \cdots, z_{n}\}$ where $z_{i}\in \mathbb{R}^{d}$. We estimate posterior probability $\mathbb{P}(Y|Z)$ and use  Eq.~\ref{eq:label} to assign each of the samples to one of $K$ cluster labels.

\begin{equation}
\label{eq:label}
\hat{k}_{x_{i}}=\underset{k}{\argmax}~\mathbb{P}(y_{k}|z)
\end{equation}

However, we reuse well-known \emph{CNN} network architecture to transform raw images into its embedded subspace. It is well-known that transfer learning expedites the process of learning new distribution. Hence, we may transfer learned weights of those networks in the beginning. 
Additionally, \emph{softmax} function is used to learn $\mathbb{P}(Y|Z)$.  Section~\ref{architecture} describes detailed architecture of our model.
\subsection{Unsupervised Training Criterion}
{\bf Motivation}: Designing a loss function for unsupervised classification is highly influenced by the concept of ``structuredness" and ``diversity" of the dataset. Data can be well predicted  if it belongs to a uniform distribution. In information theory, this ``structuredness" is quantified by low entropy. On the other hand, multi class environment is characterized by ``diversity" in data, which increases entropy of the dataset. 	
A clustering algorithm becomes vulnerable when it produces highly skewed partition. Thereby, it witnesses low entropy. Therefore, it is desirable that an algorithm should report low entropy for a partition only when it is non-skewed.\\
{\bf Deep Network Scenario}: We discuss this issue in the context of deep architecture. In general, deep networks are trained by the dataset which is divided into mini-batches. They accumulate losses incurred by each mini-batch and update the parameters of the network by incorporating the gradient of the loss function. 
We consider that a data is well classified by the network when the probability that it corresponds to a particular class out of $K$ classes, is near about 1. Therefore, it estimates low entropy. To accomplish it, classification entropy ($H_{\Phi}$) is designed. However, skewed partitioning is reduced by enforcing ``diversity" in the mini-batch. The underlying intuition is that more the number of distinct representatives the architecture experiences in a mini-batch, better it learns the grouping. To accomplish it, class entropy ($H_{\Psi}$) is designed. Therefore, the aim is to maximize the class entropy while minimizing classification entropy.\\
{\bf Mathematical Formulation}: In our model, we have feature extraction layers and classification layers. Feature extraction layers provide the mapping from input image space to feature embedded space $f_{\theta_{1}}:X \rightarrow Z$. However Classification layers provide the mapping of feature embedded space to cluster labels space, i.e.$f_{\theta_{2}}:Z \rightarrow R$. We are interested in knowing the probability that the $m^{th}$ sample belongs to $l^{th}$ class, which is denoted as $p_{ml}$. i.e.
\begin{equation}
p_{ml}=P(y_{m}=l|R)=\frac{exp({r_{ml}})}{\sum_{j=1}^{K}exp({r_{mj}})}
\end{equation}
We design our loss function ($L$) for unsupervised classification with aforesaid motivation as follows :
\begin{equation}
\label{eq:loss}
\begin{array}{lll}
L&=&\lambda H_{\Phi}-H_{\Psi}\\ \\
H_{\Phi}&=&-\frac{1}{M}\sum_{m=1}^{M}\sum_{l=1}^{K}p_{ml}\log_{}(p_{ml})\\ \\

H_{\Psi}&=&-\sum_{l=1}^{K}p_{l}\log_{}(p_{l})\\ \\

p_{l}&=&\frac{\sum_{m=1}^{M}(p_{ml})}{\sum_{l=1}^{K}\sum_{m=1}^{M}(p_{ml})}\\ \\

\end{array}
\end{equation}
 
where $\lambda$ weighs loss function $L$.

\subsection{Implementation Bottleneck} 
 In our work, mini-batch size is a prime factor for successful implementation of our proposition. It is important that network should see all possible cluster representatives in one go. To maintain this criterion, mini batch size should be sufficiently large. However, with the constrained size of {\emph GPU}, it could be hard to train network with a large batch-size. Therefore, Algorithm~\ref{algo-1} may not be reliable for assigning efficient cluster labels to the input samples. To deal with such bottleneck, we employ a strategy to increase the mini-batch size. We can store losses, incurred by $t-1$ mini-batches using only $K+1$ variable. Therefore, $t^{th}$ mini-batch will accumulate previous semi processed losses and generate loss for all the samples that network visits till now. Then, it back propagates the accumulated loss in the network. With this strategy, it is possible to execute our algorithm even in the \emph{GPU} with constrained memory. Algorithm~\ref{algo-2} depicts the strategy.

\begin{algorithm}
	\SetKwInOut{Input}{Input}
	\SetKwInOut{Output}{Output}
	
	\Input{\\X=Collection of Image Data\\		K=Target number of cluster\\ B= Batch Size}
	\Output{\\$Y^{*}$ : image  labels \\
	$\theta^{*}$ : CNN parameters}
	
	Initialize $\theta$, 
	optimizer=(\emph{SGD}, lr=0.0001)\\
	\Repeat{converged}
	{
		\ForEach{minibatch}
		{
			Compute $\mathbb{P}(Y|Z)$ by forward propagation using $\theta$.\\
			compute 
			{\em Loss= $\lambda H_{\Phi}-H_{\Psi}$} using Eq.\ref{eq:loss}.\\
				Update $\theta$ by back propagation.
		
		}

	}

	\caption{Joint optimization on $y$ and $\theta$}
	\label{algo-1}
\end{algorithm}

\begin{algorithm}
	\SetKwInOut{Input}{Input}
	\SetKwInOut{Output}{Output}
	
	\Input{\\X=Collection of Image Data\\		K=target number of cluster\\ 
		M= Number of samples in each mini-batch \\ t=number of mini-batches in one pass}
	\Output{\\$Y^{*}$ : image  labels \\
		$\theta^{*}$ : CNN parameters}
	
	Initialize $\theta$, 
	\emph{SGD}=(lr,momentum),mini-batch-size=0\\
	\Repeat{converged}
	{
		\ForEach{minibatch ($B$)}
		{
			
			mini-batch-size+=size of $B$.\\
			Compute $\mathbb{P}(Y|Z)$ by forward propagation using $\theta$.\\
			\eIf{t mini-batches visited}
			{
				
				$h_{\phi}=h_{\phi}+\sum_{m=1}^{M}\sum_{l=1}^{K}p_{ml}\log_{}(p_{ml})$\\
				$H_{\Phi}=h_{\phi}/$mini-batch-size\\
			 	$h_{\psi}=h_{\psi}+\sum_{m=1}^{M}(p_{ml})$\\
				 $p_{l}=\frac{(h_{\psi}))}{\sum_{k=1}^{K}(h_{\psi})}$\\
				 $H_{\Psi}=-\sum_{l=1}^{K}p_{l}\log_{}(p_{l})$\\
				 {\em Loss= $\lambda H_{\Phi}-H_{\Psi}$}\\
				Update $\theta$ by back propagation.\\
				mini-batch-size=0
			}  
			{
				 $h_{\phi}=h_{\phi}+\sum_{m=1}^{M}\sum_{l=1}^{K}p_{ml}\log_{}(p_{ml})$ \\
				 $h_{\psi}=h_{\psi}+\sum_{m=1}^{M}(p_{ml})$\\
				store $h_{\phi}$, $h_{\psi}.$
			}

		}

	}
	
	\caption{Joint optimization on $y$ and $\theta$ for large size of minibatch beyond the available \emph{GPU} capacity }
	\label{algo-2}
\end{algorithm}

\section{Experiments}
In this section, we first compare \emph{RECAL} with state-of-the-art clustering methods on several bench-mark image datasets. Then, we run \emph{k-means} clustering algorithm 
on our learned deep representations. Finally, we analyze the performance of \emph{RECAL} model on unsupervised segmentation task on multi-spectral \emph{LISS-IV} images.

\subsection{Alternative Clustering Models}
We compare our model with several baseline and deep models, including k-means, normalized cuts (NCuts)~\cite{N-Cut_2000}, self-tuning spectral clustering (SC-ST)~\cite{SC-ST_NIPS04},
large-scale spectral clustering (SC-LS)~\cite{SC-LS15},~ 
graph degree linkage-based agglomerative clustering (AC-GDL)~\cite{AC-GDL12}, agglomerative clustering via path integral (AC-PIC)~\cite{AC-PIC_13},
local discriminant models and global integration (LDMGI)~\cite{Local-Discriminant-Models_Global-Integration_10}, NMF with deep model (NMF-D) \cite{Deep-semi-NMF14}, task-specific clustering
with deep model (TSC-D) \cite{TSC-D15}, deep embedded clustering (DEC)\cite{DEC2015}, joint unsupervised learning (JULE)~\cite{JULE2016} and deep embedded regularized clustering (DEPICT)~\cite{DEPICT} for unsupervised clustering.
\begin{table*}[!htb]
	\center
	
	\caption{Description of Datasets}
	\begin{tabular}{lccccccc}
		\hline
		Dataset & MNIST-full & USPS  & COIL-20 & COIL-100 & U-Mist & CMU-PIE & YTF \\ \hline
		\#Samples & 70000 & 9298  & 1440 & 7200 & 575 & 2856 & 10000 \\ 
		\#Categories & 10 & 10  & 20 & 100 & 20 & 68 & 41 \\ 
		Image Size & 28$\times$ 28 & 16 $\times$ 16 &  128 $\times$ 128 & 128 $\times$ 128 & 112 $\times$ 92 & 32 $\times$ 32 & 55 $\times$ 55 \\ \hline
	\end{tabular}
	\label{tab:datasets}
\end{table*}
\subsection{Dataset}
We have chosen different types of datasets to show consistent behavior of our model towards accomplishing clustering. As, we are examining unsupervised models, we concatenate training and test datasets whenever applicable. \emph{MNIST-full}~\cite{MNIST}: It contains monochrome images of handwritten digits.~
\emph{USPS}\footnote{\url{https://www.cs.nyu.edu/~roweis/data.html}}:~It is a handwritten digits dataset from the \emph{US} postal services.
~\emph{COIL-20} and \emph{COIL-100}~\cite{COIL}: They are databases of multi-view objects. 
\emph{UMist}~\cite{UMist}: It includes face images of 20 people.~\emph{CMU-PIE}~\cite{cmu-pie}: It includes face images of 4 different facial expressions.~\emph{Youtube-Face (YTF)~\cite{YTF}}: We follow an experimental set up for this dataset as in ~\cite{JULE2016,DEPICT}. We select the first 41 subjects of \emph{YTF} dataset. Faces inside images are first cropped and then resized to 55 by 55 sizes. A brief description of these datasets is provided in Table~\ref{tab:datasets}.


\subsection{Evaluation Metric}
We follow current literature which uses normalized mutual information (NMI)~\cite{NMI} 
 as evaluation criteria for clustering algorithms. \emph{NMI} evaluates the similarity between two labels of same dataset in normalized form (0,1). However, $0$ implies no correlation and 1 shows perfect correlation between two labels. We use predicted labels (by the model) and the ground truth labels to estimate \emph{NMI}. 

\subsection{Implementation Details} \label{architecture}
We adopt feature extraction layers of \emph{VGG16}~\cite{VGGnet} with batch normalization and use the weights of the  original network trained with \emph{ImageNet} as a starting point. However, we consider a fully connected layer followed by a batch normalization and a \emph{ReLU} layer for unsupervised labeling (equivalent classification layers). We stacked classification layers  on top of feature extraction layers of \emph{VGG16}. For \emph{USPS} dataset, we decapitate feature layers of \emph{VGG16} to get an output of size $1 \times 1$, otherwise, we use all feature layers of \emph{VGG16}. Consequently, we normalize the image intensities using  Eq.~\ref{eq:normalize} where ($\mu$) and  ($\sigma$) are mean and standard deviation  of the given dataset, respectively. Moreover, we use stochastic gradient discent (SGD) as our optimization method with the learning rate=0.0001 and momentum=0.9. The weights of fully connected layer is initialized by \emph{Xavier} approach~\cite{xavier}. 
Pre-trained models may not generate useful features to obtain a good clustering for the target dataset. Hence, fine-tuning is required. Pre-trained models not only expedite the process of convergence in fine-tuning but, it helps to provide a global solution instead of a local one.
\begin{equation}
\label{eq:normalize}
\hat{x}=\frac{x-\mu}{\sigma}
\end{equation}

\subsection{Quantitative Comparison}
We show \emph{NMI} for every method on various datasets. Experimental results are averaged over 3 runs. We borrowed best results from either original paper or \cite{JULE2016} if reported on a given dataset. Otherwise, 
we put dash marks (-) without reporting any result. We report our results in two parts: 1) the clustering labels predicted by our network after training it for several epochs; 2) the clustering results obtained by running \emph{k-means} clustering algorithm on learned representation by our network.
Table~\ref{tab:NMI} reports the clustering metric, normalized mutual information \emph{(NMI)}, of the comparative algorithms along with \emph{RECAL} on the aforementioned datasets. We have transferred weights of feature layers of \emph{VGG-16} (trained with  \emph{ImageNet}) to the feature layers of \emph{RECAL} and classification layers are initialized with \emph{Xavier}'s approach. As shown, \emph{RECAL} perform competitively and consistently. It should be noted that hyper parameter tuning using labeled samples is not feasible always in real world clustering task. \emph{DEPICT} has already shown an approach of not using supervised signals. However, their approach is dependent on other clustering algorithm. Hence, our algorithm is an independent and significantly better clustering approach for handling real world large scale dataset. Interestingly, clustering results on learned representatives produce better outcome as shown in Table~\ref{tab:clusteringNMI}. It indicates that our method can learn more discriminative feature representatives compared to image intensity. Notably, our learned representation is much better than \emph{JULE}~\cite{JULE2016} for multi-viewed objects and digit dataset where their clustering result by their model, produce perfect \emph{NMI} on \emph{COIL-20}. However, our method performs poorly on face datasets. Also, Table~\ref{tab:NMI} suggests that learned features are much distinctive compared to the features obtained by {\emph VGG-16} model trained with {\emph ImageNet}.
We have executed our code on a machine with \emph{GeForce GTX 1080 Ti} GPU.

\section{Multi-spectral image segmentation}
For illustrating usefulness of this approach, we consider application of this method for unsupervised labeling of pixels in an image. We experiment with satellite  multi-spectral images. We have observed that very tiny patches may be a representative of a pure class. With this assumption, we split the pre-processed satellite  image into small patches (e.g, $32 \times 32$ or $16 \times 16$ or $8 \times 8$) with strides of $k$ (e.g,  $16$ or $8$) pixels. However, we decapitate the feature layers of  \emph{VGG-16} from its tail such that size of the feature (not output of whole model)  for an input patch of $8 \times 8$  is $1 \times 1$. Otherwise, there is no change in the network architecture. We do our experiments with other large size patches on this decapitated version of the network.
\subsection{Description of Dataset}
For our experiments, we have used images captured by Linear
Imaging Self Scanner \emph{LISS-IV} sensor of \emph{RESOURCESAT-2}. Table~\ref{tab:liss4} illustrates the specifications of \emph{LISS-IV} sensor and scene details. It includes sensors, acquisition dates and other specification
of the Indian Remote Sensing Resourcesat-2 mission. We have taken two scenes for our experiment. Scene information is shown in Table~\ref{tab:scene_info}. Both scenes are captured by the same \emph{LISS-IV} sensor on different dates.
As field investigation is a crucial task and requires expertise, we have used Microsoft\textregistered~BingTM maps for validation of our experiment.

\begin{table}[htb!]
	
	\caption{Specification of \emph{LISS-IV} sensor and scene information}
	\begin{tabular}{ll}
		\hline
		\textbf{Description of Parameter} & \textbf{Details} \\ \hline
		Agency & NRSC \\ 
		Satellite & Resourcesat 2 \\ 
		Sensor &  LISS-IV \\ 
		Swath (km) & 70 \\ 
		Resolution (m) & 5.8 \\ 
		Repetivity (days) & 5 \\ 
		Qunatisation & 10-bit \\ 
		No of Bands & 3 \\ 
		Spectral Bands ($\mu$) & \begin{tabular}[c]{@{}l@{}}Band 2 (0.52 - 0.59 )\\ Band 3 (0.62 - 0.68 )\\ Band 4 (0.77 - 0.86 ) 
		\end{tabular}\\\hline
	\end{tabular}
	\label{tab:liss4}
	
\end{table}

\begin{table}[htb!]
	\center
	
	\caption{Description of the study area}
	\begin{tabular}{lll}
	\hline
	\textbf{Scene} & \textbf{Parameters} & \textbf{Details}\\\hline
	\multirow{3}{*}{Scene-1} & Acquisition Date & 25-NOV-2011 \\
	& Scene Center Lattitude & 22.426371 N \\ 
	& Scene Center Longitude & 88.169285 E \\ \hline
	\multirow{3}{*}{Scene-2} & Acquisition Date & 10-APR-2017 \\
	&Scene Center Lattitude & 23.615814 N \\ 
	& Scene Center Longitude & 87.400460 E \\ \hline
   \end{tabular}
\label{tab:scene_info}
\end{table}
\subsection{Description of Scene}
Scene-1 covers an approximately 21 km $\times$ 23 km rectangle region located surrounding the city of Kolkata in West Bengal, India. This region mainly contains built-up areas, bare lands, vegetation (man made gardens, play grounds, small agricultural land etc), and water body. We have cropped small regions from the original data. On the other hand, Scene-2 is the whole dataset. It covers nearly 71 km $\times$ 86 km rectangle region located surrounding the place of Ranugunj, Durgapur, Panagarh, Dubrajpur etc. in West Bengal, India. However, this scene contains large hierarchy of classes. It contains built-up area, bare lands, vegetation (forest area, agricultural land, man made gardens etc.), water-body, coal mining area, etc. Moreover, built-up area covers a large area in Scene-1, whereas vegetation region predominates in Scene-2.
\subsection{Pre-processing of dataset}
In general, a sensor captures  Earth's surface radiance and this response is quantized into $n$-bit values. It is commonly called Digital Number (\emph{DN}). However, to convert calibrated (\emph{DN}s) to at-sensor radiance, two meta-parameters are used: known dynamic-range limits of the instrument and their corresponding pre-defined spectral radiance values. Meta files are provided by the satellite data provider. This radiance is then converted to top-of-atmosphere (\emph{ToA}) reflectance by normalizing for solar elevation and solar spectral irradiance.  
\subsection{Results} 
\begin{figure}
	\centering
	\includegraphics[width=0.4\textwidth, height=0.3\textheight]{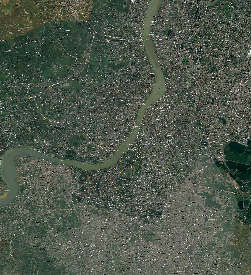}
	\caption{Scene-1, Bing Hybrid Map}
	\label{fig:bings1}
\end{figure}

\begin{figure}
	\centering
		\includegraphics[width=0.4\textwidth, height=0.3\textheight]{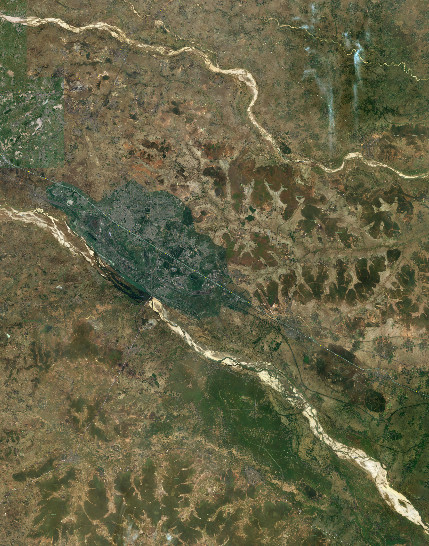}
		\caption{Scene-2, Bing Hybrid Map}
		\label{fig:bings2}
\end{figure}

In this experiment, we separately trained the model with the patches from Scene-1 and Scene-2 for $K=10$. Lets call it model-1 and model-2 respectively. We observe each segmented output from model-1 and model-2 separately. However, we check cross-dataset segmentation output which measures the robustness of the network. i.e, We use model-1 to generate the segmentation for scene-2 and vice versa. Notably, number of patches from Scene-1 and Scene-2 are approx. 0.2 million and 2.2 million respectively. Models took nearly 1-1.5 days for training. For Scene-1, our model has detected built-up area, vegetation and water-body distinctively among 10 classes. However, model-2 identifies vegetation, water body and wetland area. This is to be noted that model-2 predicts vegetation into three distinct classes. This result might be indicator of three types of vegetation in that location. We have shown models' predictions and Microsoft\textregistered~BingTM maps of the scenes in Fig~\ref{fig:scene_prediction_with_ccudata},~\ref{fig:scene_prediction_with_ranigunjdata} and Fig~\ref{fig:bings1},~\ref{fig:bings2}, respectively. Notably, Fig~\ref{fig:waterbody_s1} and Fig~\ref{fig:crosswetland_s2} show different types of region for a particular class in two scenes: built-up area in Scene-1 and wetland in Scene-2. Many of previous researches~\cite{built-up1,built-up2,built-up3} have reported similar characteristics of built-up area with wetland, farmland, etc. Similar false alarms persist in our experiment too. We use same color symbol for showing same cluster number in both scene-1 and scene-2 for the same model in Fig~\ref{fig:scene_prediction_with_ccudata}, \ref{fig:scene_prediction_with_ranigunjdata}.
\subsubsection{Analysis}
\begin{figure*}
	\centering
	\begin{tabular}[l]{lllll}
		\centering

\begin{subfigure}[b]{0.18\textwidth}
	\includegraphics[width=\textwidth]{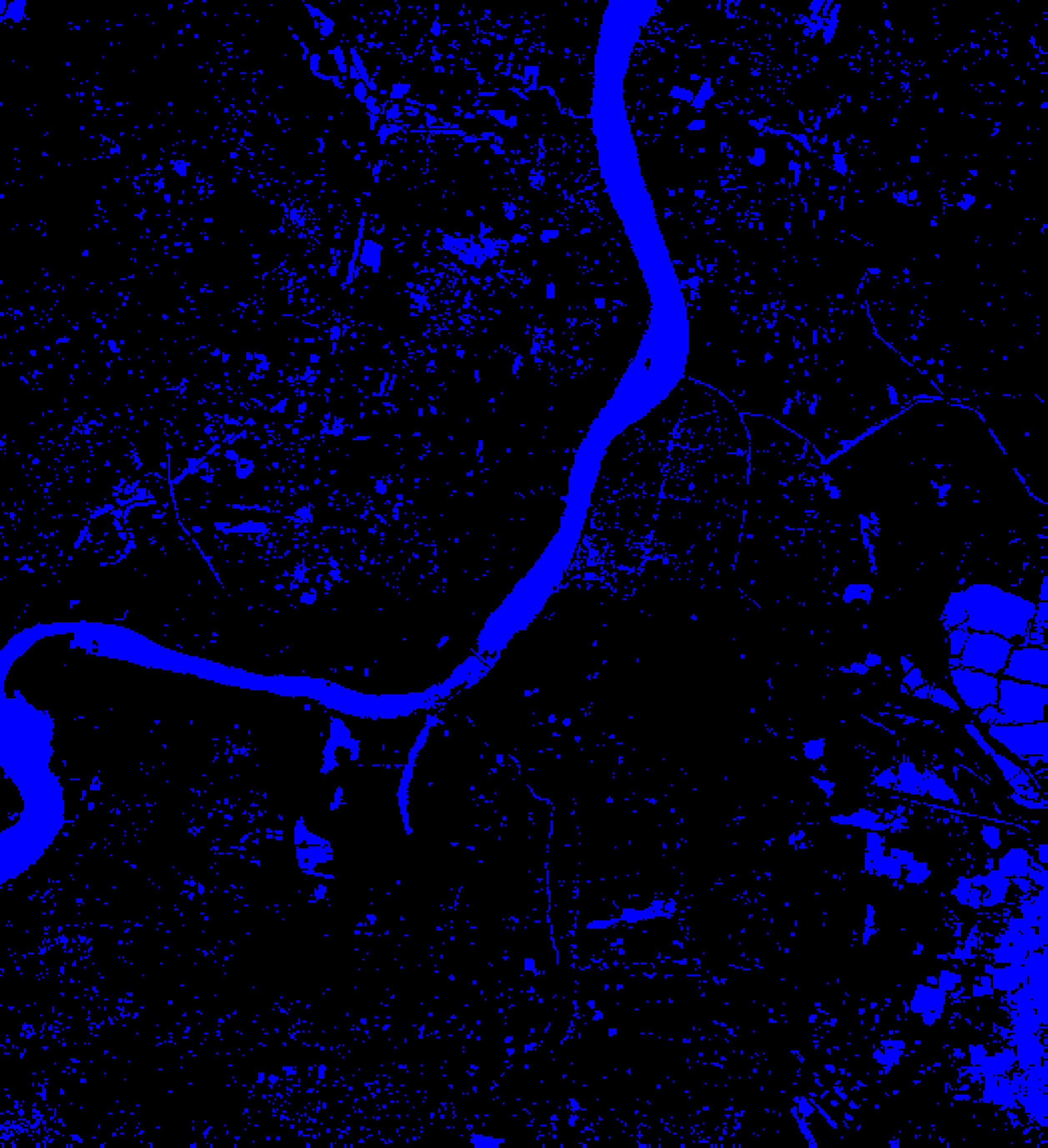}
	\caption{water body, Scene-1 \centering(WBS1)}
	\label{fig:waterbody_s1}
\end{subfigure}&
\begin{subfigure}[b]{0.18\textwidth}
	\includegraphics[width=\textwidth]{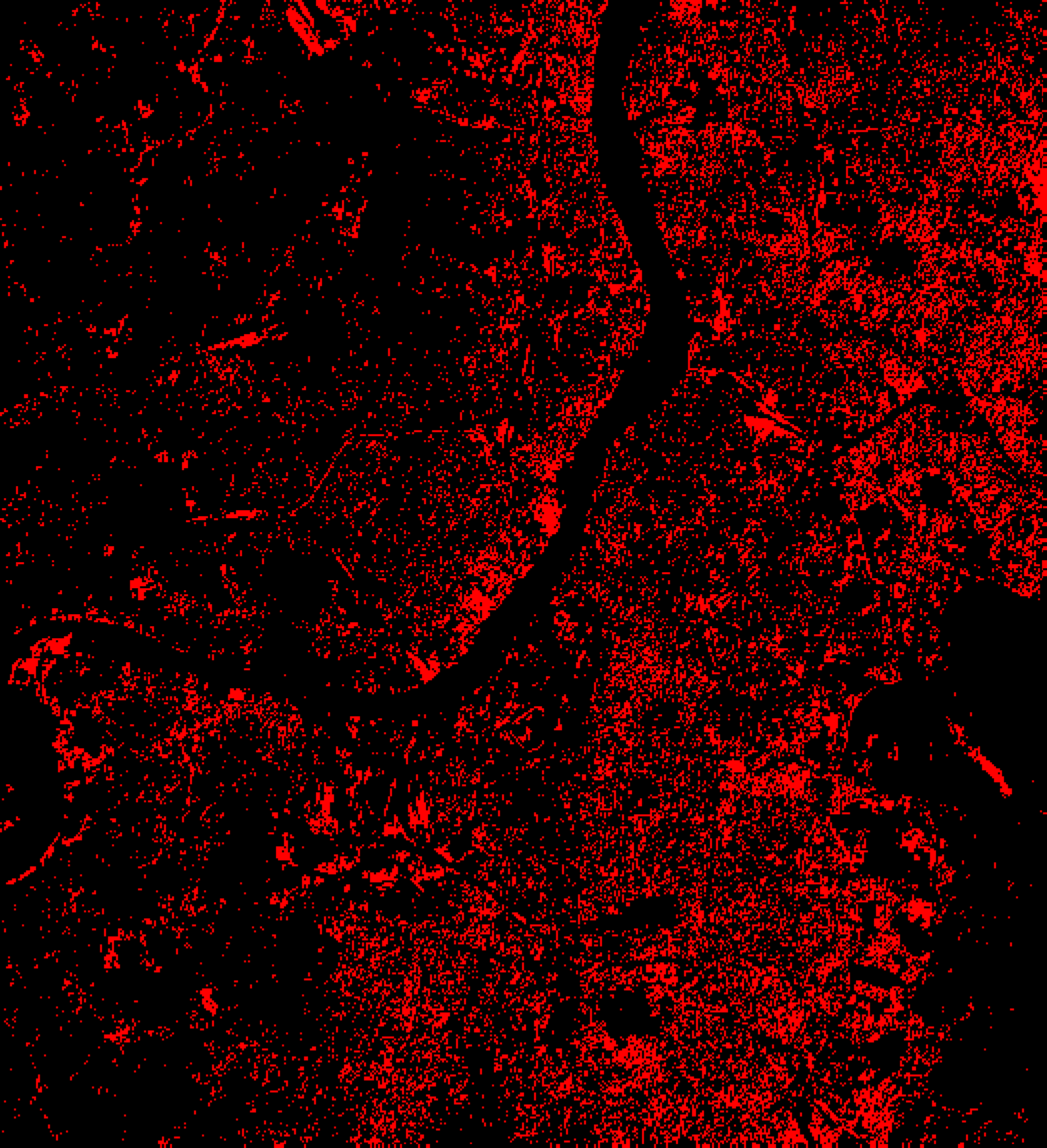}
	\caption{built-up1, Scene-1 \centering(B1S1)}
	\label{fig:built-up_s1}
\end{subfigure}&
\begin{subfigure}[b]{0.18\textwidth}
	\includegraphics[width=\textwidth]{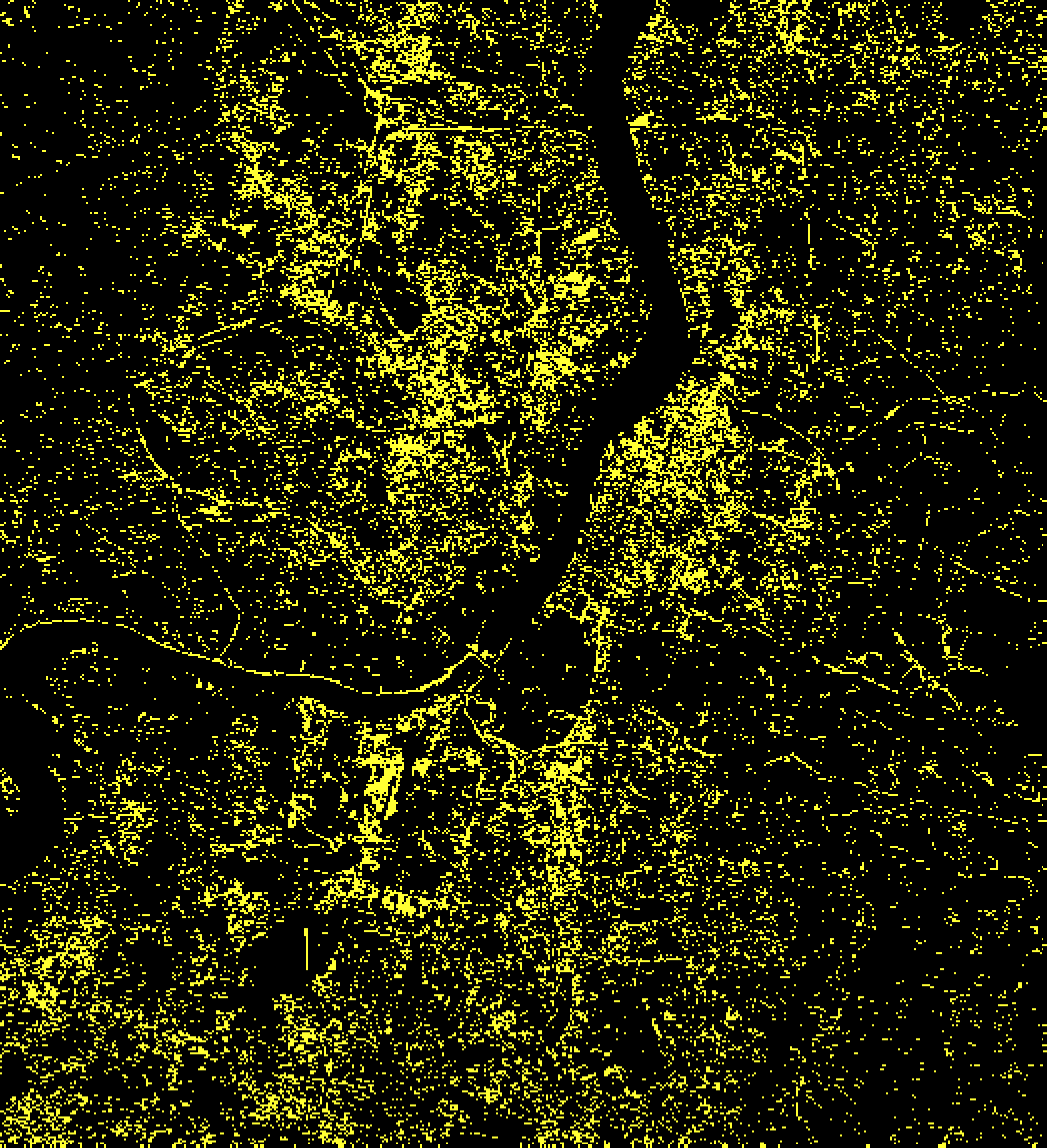}
	\caption{built-up2, Scene-1 \centering(B2S1)}
	\label{fig:veg1_s1}
\end{subfigure} &
	\begin{subfigure}[b]{0.18\textwidth}
	\includegraphics[width=\textwidth]{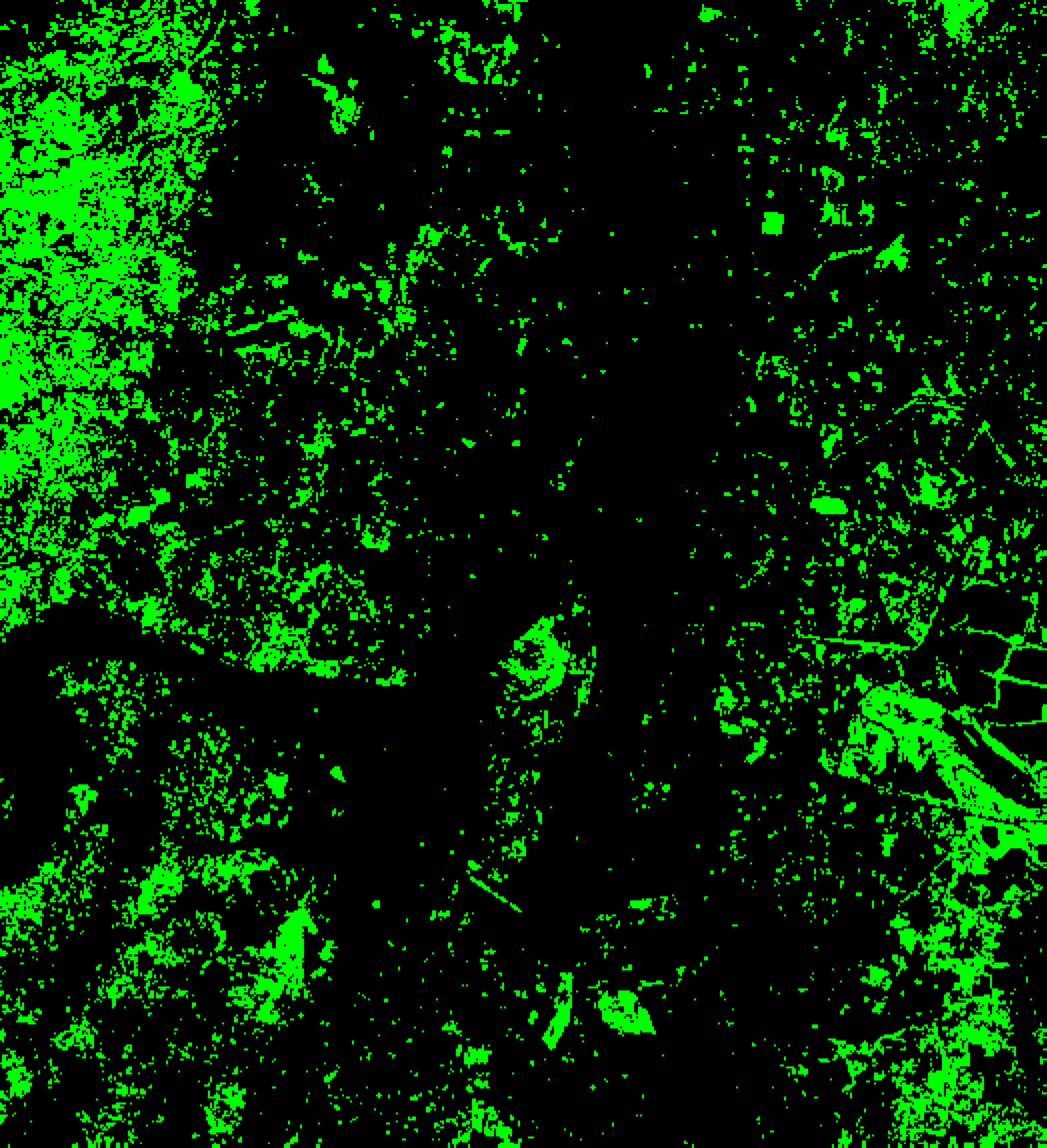}
	\caption{vegetation, Scene-1 \centering(V1S1)}
	\label{fig:veg2_s1}
\end{subfigure}&
\begin{subfigure}[b]{0.18\textwidth}
	\includegraphics[width=\textwidth]{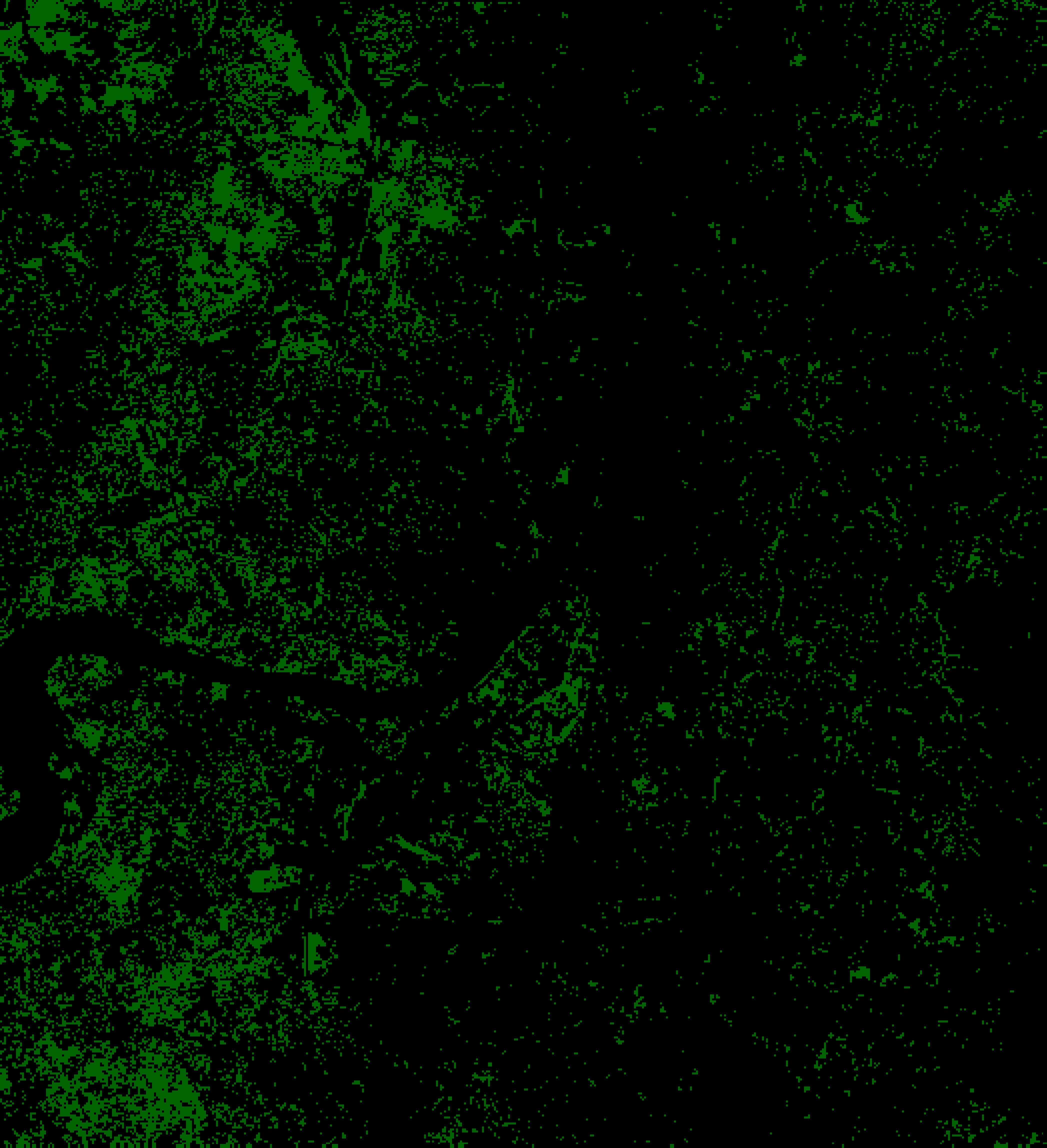}
	\caption{vegetation, Scene-1 \centering(V2S1)}
	\label{fig:veg3_s1}
\end{subfigure}
 \\

		\begin{subfigure}[b]{0.18\textwidth}
			\includegraphics[width=\textwidth]{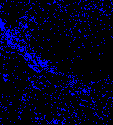}
			\caption{waterbody, Scene-2 \centering(cWBS2)}
			\label{fig:crosswaterbody_s2}
		\end{subfigure}&
		\begin{subfigure}[b]{0.18\textwidth}
			\includegraphics[width=\textwidth]{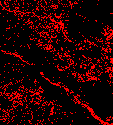}
			\caption{wet-land, Scene-2 \centering(cWLS2)}
			\label{fig:crosswetland_s2}
		\end{subfigure}&
		\begin{subfigure}[b]{0.18\textwidth}
			\includegraphics[width=\textwidth]{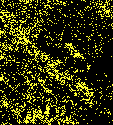}
			\caption{vegetation, Scene-2 \centering(cV1S2)}
			\label{fig:crossveg1_s2}
		\end{subfigure}&
		\begin{subfigure}[b]{0.18\textwidth}
			\includegraphics[width=\textwidth]{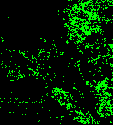}
			\caption{vegetation, Scene-2 \centering(cV2S2)}
			\label{fig:crossveg2_s2}
		\end{subfigure}&
	
		\begin{subfigure}[b]{0.18\textwidth}
			\includegraphics[width=\textwidth]{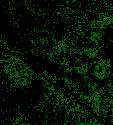}
			\caption{vegetation, Scene-2 \centering(cV3S2)}
			\label{fig:crossveg3_s2}
		\end{subfigure}
	\end{tabular}
	\caption{First and second rows show segmentation results of scene-1 and scene-2 respectively. The result set is generated using model-1. }\label{fig:scene_prediction_with_ccudata}
\end{figure*}

\begin{figure*}
	\centering
	\begin{tabular}[l]{lllll}
		\centering
		
		\begin{subfigure}[b]{0.18\textwidth}
			\includegraphics[width=\textwidth]{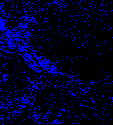}
			\caption{water body, Scene-2 \centering(WBS2)}
			\label{fig:waterbody_s2}
		\end{subfigure}&
		
		\begin{subfigure}[b]{0.18\textwidth}
			\includegraphics[width=\textwidth]{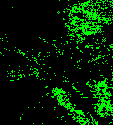}
			\caption{vegetation, Scene-2 \centering(V1S2)}
			\label{fig:vegetation2_s2}
		\end{subfigure}&
		\begin{subfigure}[b]{0.18\textwidth}
			\includegraphics[width=\textwidth]{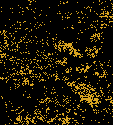}
			\caption{Vegetation, Scene-2 \centering(V2S2)}
			\label{fig:vegetation1_s2}
		\end{subfigure}&
	
	\begin{subfigure}[b]{0.18\textwidth}
		\includegraphics[width=\textwidth]{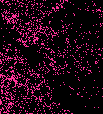}
		\caption{Vegetation, Scene-2 \centering(V3S2)}
		\label{fig:vegetation3_s2}
	\end{subfigure}&
		\begin{subfigure}[b]{0.18\textwidth}
			\includegraphics[width=\textwidth]{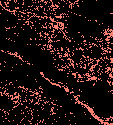}
			\caption{Wetland, Scene-2 \centering(WLS2)}
			\label{fig:wetland_s2}
		\end{subfigure}\\

		\begin{subfigure}[b]{0.18\textwidth}
			\includegraphics[width=\textwidth]{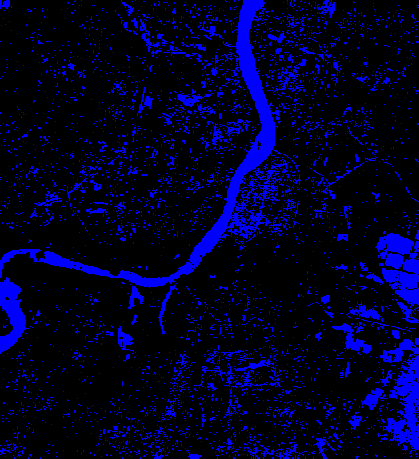}
			\caption{water-body, Scene-1 \centering(cWBS1)}
			\label{fig:water-body}
		\end{subfigure}&
		\begin{subfigure}[b]{0.18\textwidth}
			\includegraphics[width=\textwidth]{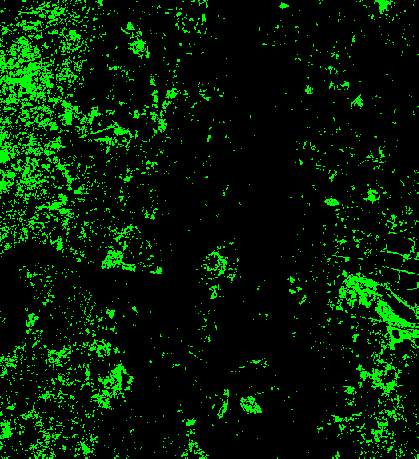}
			\caption{Vegetation, Scene-1 \centering(cV1S1)}
			\label{fig:forest1}
		\end{subfigure}&
		\begin{subfigure}[b]{0.18\textwidth}
			\includegraphics[width=\textwidth]{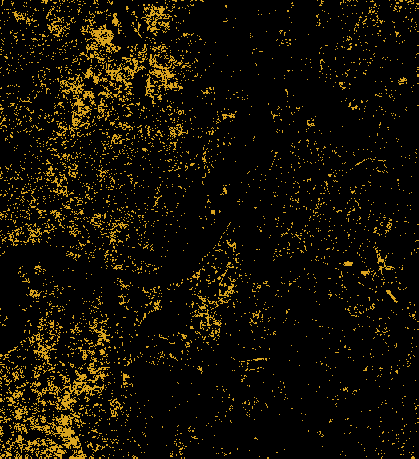}
			\caption{vegetation, Scene-1 \centering(cV2S1)}
			\label{fig:forest2}
		\end{subfigure}&
		\begin{subfigure}[b]{0.18\textwidth}
			\includegraphics[width=\textwidth]{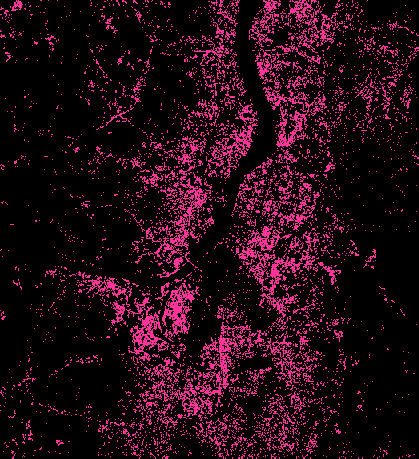}
			\caption{built-up, Scene-1 \centering(cB1S1)}
			\label{fig:builtup3}
		\end{subfigure}&
		\begin{subfigure}[b]{0.18\textwidth}
			\includegraphics[width=\textwidth]{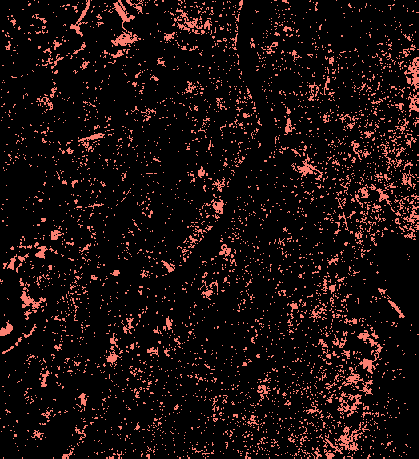}
			\caption{built-up, Scene-1 \centering(cB2S1)}
			\label{fig:wetland}
		\end{subfigure}
	\end{tabular}
	\caption{First and second rows show segmentation results of scene-2 and scene-1 respectively. The result set is generated using model-2.}
	\label{fig:scene_prediction_with_ranigunjdata}
\end{figure*}

Experimental results reflect the generalization of  model-1 and model-2 in various classes in Fig~\ref{fig:scene_prediction_with_ccudata}, \ref{fig:scene_prediction_with_ranigunjdata}. For example, Fig \ref{fig:waterbody_s1}, \ref{fig:water-body} provide visually similar structure for water-body. We prepare a validation set using  Microsoft\textregistered~BingTM maps of that region. It shows 97\% accuracy in identifying water-body in validation set. Taking visual consensus of two models, we are able to identify a few such structures for vegetation and wet land. For Scene-2 following pair of figures describe visually similar structure: (Fig~\ref{fig:crossveg1_s2}, Fig~\ref{fig:vegetation1_s2}), (Fig~\ref{fig:crossveg2_s2}, Fig~\ref{fig:vegetation2_s2}) and (Fig~\ref{fig:crosswetland_s2}, Fig~\ref{fig:wetland_s2}). Similarly, (Fig~\ref{fig:veg2_s1}, Fig~\ref{fig:forest1}) and (Fig~\ref{fig:veg3_s1}, Fig~\ref{fig:forest2}) reveal similar visual structure in Scene-1. We measure consensus of two models using {\em Jaccard similarity} score as shown in  Eq~\ref{eq:consensus}. 
\begin{equation}
\label{eq:consensus}
\text{\emph{ Jaccard similarity}}=\frac{\abs{M1 \cap M2}}{\abs{M1 \cup M2}}
\end{equation}
where $M1$ and $M2$ represent pixels predicted by model-1 and model-2, respectively. 
Table-\ref{tab:confusion_matrix_s1}  and Table-\ref{tab:confusion_matrix_s2} show the \emph{Jaccard similarity} score for various identified classes in scene-1 and scene-2, respectively. We have shown upper triangular matrix in Table~\ref{tab:confusion_matrix_s1},~\ref{tab:confusion_matrix_s2} due to the symmetric nature of this score. We follow similar naming convention for denoting a particular class for a specific scene as shown in Fig~\ref{fig:scene_prediction_with_ccudata}, \ref{fig:scene_prediction_with_ranigunjdata}.
However, further study is required for validating different classes with ground truth data, such as vegetation, wet land, built up areas, etc. 

\begin{table}[htb!]
	\centering
	\caption{{\em Jaccard similarity} score for different classes (as shown in Fig~\ref{fig:scene_prediction_with_ccudata}, \ref{fig:scene_prediction_with_ranigunjdata}) for scene-1}
	\begin{tabular}{ll|lllll|}
		\cline{3-7}
		&  & \multicolumn{5}{c|}{Model-2} \\ \cline{3-7} 
		&  & cV1S1 & cV2S1 &cB2S1 &cB1S1 & cWBS1\\ \hline
		\multicolumn{1}{|c|}{\multirow{5}{*}{\begin{sideways}Model-1\end{sideways}}} & V1S1 & \textbf{0.545} & 0.004 & 0.004& 0.002 &0.004\\ 
		\multicolumn{1}{|l|}{} & V2S1 &  &\textbf{0.299} & 0.015& 0.014&0.001 \\ 
		\multicolumn{1}{|l|}{} & B1S1 & &  &\textbf{0.243} & 0.105&0.007\\
		\multicolumn{1}{|l|}{} & B2S1 &  & &  & \textbf{0.211} & 0.004\\
		\multicolumn{1}{|l|}{} & WBS1 &  & & & & \textbf{0.721} \\ \hline
	\end{tabular}
	\label{tab:confusion_matrix_s1}
\end{table}

\begin{table}[htb!]
	\centering
	\caption{{\em Jaccard similarity} score for different classes (as shown in Fig~\ref{fig:scene_prediction_with_ccudata}, \ref{fig:scene_prediction_with_ranigunjdata}) for scene-2}
	\begin{tabular}{ll|lllll|}
		\cline{3-7}
		&  & \multicolumn{5}{c|}{Model-1} \\ \cline{3-7} 
		&  & cV2S2 & cV1S2 &cV3S2 &cWBS2 & cWLS2 \\ \hline
		\multicolumn{1}{|c|}{\multirow{5}{*}{\begin{sideways}Model-2\end{sideways}}} & V1S2 & \textbf{0.624} & 0.004&0.000 & 0.008 &0.000\\ 
		\multicolumn{1}{|l|}{} & V2S2 & & \textbf{0.186}&0.001 &0.059 & 0.000\\ 
		\multicolumn{1}{|l|}{} & V3S2 & &  & \textbf{0.315}&0.002&0.031\\
		\multicolumn{1}{|l|}{} & WBS2 &  & & & \textbf{0.556} & 0.011\\
		\multicolumn{1}{|l|}{} & WLS2 &  & & & & \textbf{0.570}\\ \hline
	\end{tabular}
	\label{tab:confusion_matrix_s2}
\end{table}

%

\begin{table*}[!htb]
	\centering
	
	\caption{Clustering performance of different algorithms on image datasets based on normalized mutual information (NMI).}
	
	\begin{tabular}{lccccccc}
		\hline
		Dataset & MNIST-full & USPS &  COIL-20 & COIL-100 & U-Mist & CMU-PIE & YTF \\ \hline
		k-means & 0.500 & 0.447   & 0.775 & 0.822  & 0.609 & 0.549 & 0.761 \\ 
		N-cuts & 0.411 & 0.675   & 0.884 & 0.861 & 0.782 & 0.411 & 0.742 \\ 
		SC-ST & 0.416 & 0.726 & 0.895 &0.858 & 0.611 & 0.581 & 0.620 \\
		SC\_LS & 0.706 & 0.681   & 0.877 & 0.833 & 0.810 & 0.788 & 0.759 \\
		AC-GDL & 0.844 & 0.824   & 0.937 & 0.933 & 0.755 & 0.934 & 0.622 \\ 
		AC-PIC & 0.940 & 0.825   & 0.950  & 0.964 & 0.750 & 0.902 & 0.697 \\
		LDMGI	&0.802 &0.563  & - & - & -& -& -\\
		 \hline\hline
		NMF-D & 0.152 & 0.287  & 0.648 & 0.748 & 0.467 & 0.920 & 0.562 \\
		TSC-D & 0.651 & - & - &  -  & - & - & - \\
		DEC & 0.816 & 0.586 & - & - & - & 0.924 & 0.446 \\
		JULE-SF & 0.906 & 0.858 & 1.000 & 0.978 & 0.880 & 0.984 & 0.848 \\
		JULE-RC & 0.913 & 0.913 & 1.000 & 0.985 & 0.877 & 1.000 & 0.848 \\
		DEPICT & 0.917 & 0.927 & - & - & - & 0.974 & 0.802 \\\hline \hline
		RECAL & 0.852 & 0.913 & 0.880 & 0.831 & 0.694 & 0.627 & 0.779 \\\hline
	\end{tabular}
	\label{tab:NMI}
\end{table*} 

\begin{table*}[!htb]
	\centering
	
	\caption{Clustering performance (\emph{NMI}) for \emph{k-means} algorithms using learned representations by different models as inputs}
	\begin{tabular}{llccccccc}
		\hline
		\multicolumn{1}{l}{Method} & Model & MNIST-full & USPS & COIL-20 & COIL-100 & UMist & CMU\_PIE & YTF \\\hline
		\multirow{3}{*}{k-means} & \emph{RECAL} & 0.885 & \textbf{0.913} & \textbf{0.948} & \textbf{0.919} &0.675  & 0.715 & 0.809 \\
		& \emph{JULE} & \textbf{0.927} & 0.758 & 0.926 & \textbf{0.919} & \textbf{0.871} & \textbf{0.956} & \textbf{0.835}\\
		& \emph{VGG-16} & 0.182 &0.014  & 0.769 &0.792  &0.576  &0.302  &0.167 \\\hline
		
	\end{tabular}
	\label{tab:clusteringNMI}
\end{table*}
%
%

\section{Conclusion}
In this paper, we have proposed an approach to transform supervised \emph{CNN} classifier architecture into an unsupervised clustering model. We address ``structuredness" and ``diversity" of the dataset to migrate from supervised to unsupervised paradigm. Our model  jointly learns discriminative embedding subspace and  cluster labels. The major strength of this approach is : scalability and re-usability. Experimental results show that \emph{RECAL} performs competitively for real-world clustering task. However, it learns representation better compared to the existing approaches for non-face image datasets. 
We have shown segmentation as one of the application by our proposed method in remote sensing domain. To achieve this, we split the original image into small patches and get label for each of them. Then, they are combined to get the segmentation of the whole image. We qualitatively observe that this model is able to detect vegetation, water-body, wet-land, built-up area separately. We also provide the power of transferring learned representation across separate scenes in this work. However, built-up area has confusion with wetland and vegetation in cross-dataset segmentation task. It is probably due to the similar characteristics among these classes which is an innate problem in the underlying dataset. We also observe that training of the proposed network is fast. It took about roughly 30 hours to process more than 2 million patches in our experiments in \emph{GeForce GTX 1080 Ti GPU}. This strongly indicates the property of scalability for our model. 




{\small
\bibliographystyle{ieee}
\bibliography{egbib}

\begin{thebibliography}{10}\itemsep=-1pt

\bibitem{DCAE_clustering}
A.~{Alqahtani}, X.~{Xie}, J.~{Deng}, and M.~W. {Jones}.
\newblock A deep convolutional auto-encoder with embedded clustering.
\newblock In {\em 2018 25th IEEE International Conference on Image Processing
  (ICIP)}, pages 4058--4062, 2018.

\bibitem{SC-LS15}
D.~{Cai} and X.~{Chen}.
\newblock Large scale spectral clustering via landmark-based sparse
  representation.
\newblock {\em IEEE Transactions on Cybernetics}, 45:1669--1680, 2015.

\bibitem{DEPICT}
K.~G. Dizaji, A.~Herandi, and H.~Huang.
\newblock Deep clustering via joint convolutional autoencoder embedding and
  relative entropy minimization.
\newblock {\em 2017 IEEE International Conference on Computer Vision (ICCV)},
  pages 5747--5756, 2017.

\bibitem{xavier}
X.~Glorot and Y.~Bengio.
\newblock Understanding the difficulty of training deep feedforward neural
  networks.
\newblock In {\em Proceedings of the Thirteenth International Conference on
  Artificial Intelligence and Statistics}, volume~9 of {\em Proceedings of
  Machine Learning Research}, pages 249--256, 2010.

\bibitem{UMist}
D.~B. Graham and N.~M. Allinson.
\newblock {\em Characterising Virtual Eigensignatures for General Purpose Face
  Recognition}, pages 446--456.
\newblock 1998.

\bibitem{resnet}
K.~He, X.~Zhang, S.~Ren, and J.~Sun.
\newblock Deep residual learning for image recognition.
\newblock {\em CoRR}, abs/1512.03385, 2015.

\bibitem{NMI}
Y.~Horibe.
\newblock Entropy and correlation.
\newblock {\em IEEE Transactions on Systems, Man, and Cybernetics},
  SMC-15(5):641--642, 1985.

\bibitem{CCNN2018}
C.~Hsu and C.~Lin.
\newblock Cnn-based joint clustering and representation learning with feature
  drift compensation for large-scale image data.
\newblock {\em IEEE Transactions on Multimedia}, 20(2):421--429, 2018.

\bibitem{denseNet}
G.~Huang, Z.~Liu, and K.~Q. Weinberger.
\newblock Densely connected convolutional networks.
\newblock {\em CoRR}, abs/1608.06993, 2016.

\bibitem{Alexnet}
A.~Krizhevsky, I.~Sutskever, and G.~E. Hinton.
\newblock Imagenet classification with deep convolutional neural networks.
\newblock In {\em Proceedings of the 25th International Conference on Neural
  Information Processing Systems - Volume 1}, NIPS'12, pages 1097--1105, 2012.

\bibitem{MNIST}
Y.~Lecun, L.~Bottou, Y.~Bengio, and P.~Haffner.
\newblock Gradient-based learning applied to document recognition.
\newblock {\em Proceedings of the IEEE}, 86(11):2278--2324, 1998.

\bibitem{COIL}
S.~A. Nene, S.~K. Nayar, and H.~Murase.
\newblock Columbia object image library (coil-20.
\newblock Technical report, 1996.

\bibitem{built-up1}
Y.~Qin, X.~Xiao, J.~Dong, B.~Chen, F.~Liu, G.~Zhang, Y.~Zhang, J.~Wang, and
  X.~Wu.
\newblock Quantifying annual changes in built-up area in complex urban-rural
  landscapes from analyses of palsar and landsat images.
\newblock {\em ISPRS Journal of Photogrammetry and Remote Sensing}, 124:89 --
  105, 2017.

\bibitem{N-Cut_2000}
J.~Shi and J.~{Malik}.
\newblock Normalized cuts and image segmentation.
\newblock {\em IEEE Transactions on Pattern Analysis and Machine Intelligence},
  22(8):888--905, 2000.

\bibitem{cmu-pie}
T.~Sim, S.~Baker, and M.~Bsat.
\newblock The cmu pose, illumination, and expression (pie) database.
\newblock In {\em Proceedings of the Fifth IEEE International Conference on
  Automatic Face and Gesture Recognition}, FGR '02, pages 53--, 2002.

\bibitem{VGGnet}
K.~Simonyan and A.~Zisserman.
\newblock Very deep convolutional networks for large-scale image recognition.
\newblock {\em CoRR}, abs/1409.1556, 2014.

\bibitem{built-up2}
Y.~{Tan}, F.~{Ren}, and S.~{Xiong}.
\newblock Automatic extraction of built-up area based on deep convolution
  neural network.
\newblock In {\em 2017 IEEE International Geoscience and Remote Sensing
  Symposium (IGARSS)}, pages 3333--3336, 2017.

\bibitem{built-up3}
Y.~{Tan}, S.~{Xiong}, and Y.~{Li}.
\newblock Automatic extraction of built-up areas from panchromatic and
  multispectral remote sensing images using double-stream deep convolutional
  neural networks.
\newblock {\em IEEE Journal of Selected Topics in Applied Earth Observations
  and Remote Sensing}, 11(11):3988--4004, 2018.

\bibitem{Deep-semi-NMF14}
G.~Trigeorgis, K.~Bousmalis, S.~Zafeiriou, and B.~W. Schuller.
\newblock A deep semi-nmf model for learning hidden representations.
\newblock In {\em Proceedings of the 31st International Conference on
  International Conference on Machine Learning - Volume 32}, ICML'14, pages
  II--1692--II--1700. JMLR.org, 2014.

\bibitem{TSC-D15}
Z.~Wang, S.~Chang, J.~Zhou, and T.~S. Huang.
\newblock Learning {A} task-specific deep architecture for clustering.
\newblock {\em CoRR}, abs/1509.00151, 2015.

\bibitem{YTF}
L.~Wolf, T.~Hassner, and I.~Maoz.
\newblock Face recognition in unconstrained videos with matched background
  similarity.
\newblock In {\em Proceedings of the 2011 IEEE Conference on Computer Vision
  and Pattern Recognition}, CVPR '11, pages 529--534, 2011.

\bibitem{DEC2015}
J.~Xie, R.~B. Girshick, and A.~Farhadi.
\newblock Unsupervised deep embedding for clustering analysis.
\newblock {\em CoRR}, abs/1511.06335, 2015.

\bibitem{JULE2016}
J.~Yang, D.~Parikh, and D.~Batra.
\newblock Joint unsupervised learning of deep representations and image
  clusters.
\newblock In {\em 2016 IEEE Conference on Computer Vision and Pattern
  Recognition (CVPR)}, pages 5147--5156, 2016.

\bibitem{Local-Discriminant-Models_Global-Integration_10}
Y.~{Yang}, D.~{Xu}, F.~{Nie}, S.~{Yan}, and Y.~{Zhuang}.
\newblock Image clustering using local discriminant models and global
  integration.
\newblock {\em IEEE Transactions on Image Processing}, 19(10):2761--2773, 2010.

\bibitem{SC-ST_NIPS04}
L.~Zelnik-Manor and P.~Perona.
\newblock Self-tuning spectral clustering.
\newblock In {\em Proceedings of the 17th International Conference on Neural
  Information Processing Systems}, NIPS'04, pages 1601--1608, 2004.

\bibitem{AC-GDL12}
W.~Zhang, X.~Wang, D.~Zhao, and X.~Tang.
\newblock Graph degree linkage: Agglomerative clustering on a directed graph.
\newblock {\em CoRR}, abs/1208.5092, 2012.

\bibitem{AC-PIC_13}
W.~Zhang, D.~Zhao, and X.~Wang.
\newblock Agglomerative clustering via maximum incremental path integral.
\newblock {\em Pattern Recognition}, 46(11):3056 -- 3065, 2013.

\end{thebibliography}
}

\end{document}